\documentclass[conference]{IEEEtran}
\IEEEoverridecommandlockouts
\usepackage{cite}
\usepackage{amsmath,amssymb,amsfonts}
\usepackage{algorithm}
\usepackage{algpseudocode}
\usepackage{graphicx}
\usepackage{textcomp}
\usepackage[table]{xcolor}
\usepackage{booktabs}
\usepackage{subcaption}
\usepackage{caption}
\usepackage{float}
\usepackage{lipsum} 
\usepackage{hyperref} 
\usepackage{tikz}

\newcommand\copyrighttext{%
  \footnotesize \textcopyright \the\year{} IEEE. Personal use of this material is permitted. Permission from IEEE must be obtained for all other uses, including reprinting/republishing this material for advertising or promotional purposes, collecting new collected works for resale or redistribution to servers or lists, or reuse of any copyrighted component of this work in other works.}
  
\newcommand\copyrightnotice{%
\begin{tikzpicture}[remember picture,overlay]
\node[anchor=south,yshift=10pt] at (current page.south) {\fbox{\parbox{\dimexpr0.75\textwidth-\fboxsep-\fboxrule\relax}{\copyrighttext}}};
\end{tikzpicture}%
}

\graphicspath{ {./images/} }
\def\BibTeX{{\rm B\kern-.05em{\sc i\kern-.025em b}\kern-.08em
    T\kern-.1667em\lower.7ex\hbox{E}\kern-.125emX}}
\begin{document}

\title{K-GBS3FCM - KNN Graph-Based Safe Semi-Supervised Fuzzy C-Means 
}

\bibliographystyle{./IEEEtran}

\author{\IEEEauthorblockN{ Gabriel Machado Santos}
\IEEEauthorblockA{\textit{Federal University of Uberlândia} \\
\textit{Computer Science Faculty}\\
São Paulo, Brazil \\
gabrielmsantos@gmail.com}
\and
\IEEEauthorblockN{Rita Maria Silva Julia}
\IEEEauthorblockA{\textit{Federal University of Uberlândia} \\
\textit{Computer Science Faculty}\\
Uberlândia, Brazil \\
rita@ufu.br}
\and
\IEEEauthorblockN{Marcelo Zanchetta do Nascimento}
\IEEEauthorblockA{\textit{Federal University of Uberlândia} \\
\textit{Computer Science Faculty}\\
Uberlândia, Brazil \\
marcelo.nascimento@ufu.br}
}

\maketitle

\copyrightnotice

\begin{abstract}
Clustering data using prior domain knowledge, starting from a partially labeled set, has recently been widely investigated. Often referred to as semi-supervised clustering, this approach leverages labeled data to enhance clustering accuracy. To maximize algorithm performance, it is crucial to ensure the safety of this prior knowledge. Methods addressing this concern are termed safe semi-supervised clustering (S3C) algorithms. This paper introduces the KNN graph-based safety-aware semi-supervised fuzzy c-means algorithm (K-GBS3FCM), which dynamically assesses neighborhood relationships between labeled and unlabeled data using the K-Nearest Neighbors (KNN) algorithm. This approach aims to optimize the use of labeled data while minimizing the adverse effects of incorrect labels. Additionally, it is proposed a mechanism that adjusts the influence of labeled data on unlabeled ones through regularization parameters and the average safety degree. Experimental results on multiple benchmark datasets demonstrate that the graph-based approach effectively leverages prior knowledge to enhance clustering accuracy. The proposed method was significantly superior in 64\% of the 56 test configurations, obtaining higher levels of clustering accuracy when compared to other semi-supervised and traditional unsupervised methods. This research highlights the potential of integrating graph-based approaches, such as KNN, with established techniques to develop advanced clustering algorithms, offering significant applications in fields that rely on both labeled and unlabeled data for more effective clustering.
\end{abstract}


\section{Introduction}
Clustering is an essential tool for uncovering new information and patterns in data \cite{xu2015comprehensive}, commonly associated with unsupervised learning, as it does not require prior knowledge for execution. Traditional unsupervised clustering algorithms, such as k-means \cite{Kmeans}, Fuzzy C-Means (FCM) \cite{FCM}, and Gaussian mixture models (GMM) \cite{GMM}, have been widely utilized across various industries.

To achieve better outcomes and improve accuracy, industries increasingly employ techniques to enhance prior and domain knowledge of the data. Researchers have developed numerous semi-supervised clustering (SSC) algorithms \cite{gan2013using} to leverage this prior knowledge effectively \cite{bair2013semi} \cite{grira2004unsupervised}. Among them, it highlights the Semi-Supervised FCM (SSFCM) \cite{pedrycz1997fuzzy}, which enhances clustering by integrating labeled and unlabeled data into the traditional FCM algorithm. 

Noteworthy here is the fact that these semi-supervised clustering methods (such as the one used in SSFCM) aim to improve the quality of pre-associated knowledge with clustered data, making data labeling a core part of the machine learning (ML) process.

Data can be labeled through methods like hand-labeling, programmatic labeling, or natural labeling. Most real-world datasets combine these approaches. The challenge in clustering is designing algorithms that extract relationships from labeled, unlabeled, and mislabeled data, integrating insights from supervised, semi-supervised, and weakly-supervised learning

To address the challenge of assessing the quality and confidence of labeled data in semi-supervised learning, researchers have introduced the concept of safety degree \cite{Wang2013SafetyAwareSC} \cite{gan2019safe}. This measure evaluates the reliability of labeled data points in the clustering process and balances the contribution of labeled and unlabeled data, ensuring effective leveraging of both.

Safety awareness in semi-supervised learning was first introduced in 2013 to implement a safety-control mechanism for safe semi-supervised classification \cite{Wang2013SafetyAwareSC}. Since then, numerous works have expanded on this idea, creating the concept of safe semi-supervised clustering (S3C) algorithms. 

For instance, in 2018, researchers proposed LHC-3FCM \cite{GAN2018384}, based on FCM, to handle mislabeled data through local homogeneous consistency. In the same year, researchers used a concept of multiple kernels to improve clustering performance for semi-supervised fuzzy algorithms (SMKFCM)\cite{SMKFCM}.

In 2019, an extension of particle swarm optimization (PSO) \cite{PSOTuto} to semi-supervised learning introduced a confidence-weighted safe method to FCM (CS3FCM) \cite{GAN2019107}, which assumes that different instances should have varying levels of confidence during the clustering process.

A few years later, in 2021, two major algorithms were introduced. One,  introduced a method for using entropy and relative entropy measures to improve clustering accuracy using labeled and unlabeled data (SMKFC-ER)\cite{SALEHI2021667}. The other, namely correntropy-based semi-supervised non-negative matrix factorization (CSNMF) \cite{PENG2021107683}, presented a robust semi-supervised non-negative matrix factorization (NMF) method designed to improve image clustering performance by effectively handling noisy data.

More recently, in 2023, Gan. et. al. have developed a novel and interesting S3C method called Adaptive safety-aware semi-supervised clustering (AS3FCM) \cite{gan2023adaptive} that improves clustering accuracy by adaptively estimating safety degrees to handle mislabeled instances. The experimental results of AS3FCM demonstrated that this adaptive approach to computing safety degrees can result in better accuracy in some cases.

Despite the advent of numerous S3C methods, the existing algorithms often struggle to capture the intrinsic relationship between labeled and unlabeled data, which limits improvements in clustering accuracy and reduces the ability to mitigate the impact of mislabeled data. Traditional methods often disregard the spatial relationship between data points, which compromises their efficiency in extracting valuable information concerning the dataset and leads to sub-optimal performance.

Therefore, the primary goal of this work is to build upon recent advancements in research on safety for semi-supervised clustering, particularly AS3FCM \cite{gan2023adaptive}, while drawing inspiration from other interesting clustering graph-based approaches \cite{Wang2023GCFlowAG} \cite{kulis2005semi}.

In AS3FCM, in order to assess the consistency of the labeled data, the authors assume a constant value for the number \textit{p} of nearest unlabeled neighbors (\textit{p}UN) - more specifically, equal to 5 -, which prevents the algorithm from adapting itself realistically and dynamically to the nature of the data during the execution of the density estimation process. Further, the confidence parameter is adaptively computed by a simplification resulting from a second order Taylor expansion and, then, solved as a standard quadratic programming optimization problem (OP). Despite being an elegant way to compute safety, it might introduce unnecessary complexity and not fully capture the intrinsic relationship between data points in a graph-based approach.

This way, the present paper aims to address these limitations by proposing a novel algorithm, named K-GBS3FCM, that leverages the connectivity between data points to compute the confidence level of labeled data and activate its impact on unlabeled data in non-linear datasets. The motivation, in this case, lies in the fact that, depending on the estimated density, a different number of unlabeled data points can be used to hone the process of evaluating the consistency of the labeled data. This means that labeled points with a higher estimated density can use fewer neighbors than data points with a lower estimated density. Also, if the average label consistency is considered strong (greater than some threshold), labeled data should have a larger influence on unlabeled data nearby to improve clustering accuracy. 

This process begins by dynamically computing \textit{p}UN using the k-nearest neighbor algorithm \cite{fix1985discriminatory} to determine the average distance of the k-nearest neighbors, which serves as a proxy for density estimation. Labels with low average distances have high estimated densities and, therefore, should have a smaller \textit{p}UN compared to labels with low estimated densities. Next, local consistency is assessed by comparing the consistency of the \textit{p}UN unlabeled neighbors for each labeled data point. Ultimately, if the average consistency of the labels exceeds a certain threshold, an additional weight is activated to increase the weight of the unlabeled portion in the cost function equation. Experimental results show that the K-GBS3FCM can capture intrinsic information on the relationship of data points by dynamically estimating neighborhood resulting in increased accuracy for some datasets. 

Therefore, the contributions of this work can be summarized as follows: 
\begin{itemize}
    \item Dynamically establishing the \textit{p}UN  nearest unlabeled neighbors for labeled data points using the KNN method.
    \item Simplifying safety degree estimation by computing the local consistency of each labeled data compared to its \textit{p}UN neighbors. 
    \item Reinforcing the influence of labeled regularization term on the unlabeled portion of the equation when safety degree exceeds a certain threshold.
\end{itemize}

The next sections are structured as follows: Section 2 provides a background on KNN, FCM, SSFCM, and the related work AS3FCM. The motivation of the authors and the details of implementation for this work will be described in Section 3. Section 4 reports the experiments and results. Section 5 analyzes the performance of the experimental databases. Finally, Section 6 presents the conclusion, discusses limitations, and outlines future work.

\section{Background and Related Work}

\subsection{KNN}
K-nearest neighbors (KNN) \cite{cover1967nearest} is a simple, non-parametric algorithm widely used in classification and regression tasks. It identifies the $K$ nearest data points based on a distance metric, often Euclidean distance. KNN has numerous applications, including text classification, image recognition, and medical diagnosis. In clustering, it serves as a baseline for more advanced methods \cite{chen2020fast}. However, KNN can be computationally intensive with large datasets, and its performance depends on the choice of $K$ \cite{abu2019effects}, where small values may cause overfitting and large values may lead to underfitting.

\subsection{FCM}
Fuzzy C-Means (FCM) is one of the most well-known algorithms in data clustering. Unlike K-Means, FCM allows data points to belong to multiple clusters with varying degrees of membership. The algorithm employs fuzzy logic to calculate the membership degree for each data point regarding a fixed number $c$ of clusters. This is achieved by minimizing an objective function, represented in \eqref{eq:1}, that measures the distance between data points and cluster centers, weighted by membership values. The algorithm iteratively updates the cluster centers and membership values until convergence. Thus, FCM computes memberships more flexibly by considering the relationship between data points and all clusters. Due to its nature, FCM typically performs better than K-Means  \cite{bora2017performance}.

\begin{equation}
\label{eq:1}
\begin{aligned}
& J_{FCM}(V, U) = \sum_{i=1}^{c} \sum_{k=1}^{n} (u_{ik})^m d_{ik}^2 \quad \\
& \text{s.t.} \quad \sum_{i=1}^{c} u_{ik} = 1, \quad \forall k = 1, \ldots, n \\
& \quad \quad 0 \leq u_{ik} \leq 1, \quad \forall i = 1, \ldots, c, k = 1, \ldots, n
\end{aligned}
\end{equation}

where $n$ is the number of data points in an unlabeled dataset X = [$x_1$, $x_2$, ..., $x_n$] $\subset \mathbb{R}^{dim \times n}$, \textit{dim} represents the number of features, in other words, data dimensionality, $ c$ is a pre-defined fixed number of clusters. $m$ is the fuzziness parameter usually kept as 2 and $d$ is the distance, typically the Euclidean distance, between data point $k$ and the center of cluster $v_i$. Therefore, $u_{ik}$ represents the membership value of data point $k$ to cluster $i$. The membership and cluster center are iteratively updated as shown in \eqref{eq:uik_fcm} and \eqref{eq:vi_fcm}, respectively.

\begin{equation}\label{eq:uik_fcm}
\begin{aligned}
u_{ik} = \frac{1}{\sum_{j=1}^{c} \left( \frac{d_{ik}}{d_{jk}} \right)^{\frac{2}{m-1}}}, \quad \forall i = 1, \ldots, c, k = 1, \ldots, n
\end{aligned}
\end{equation}

\begin{equation}\label{eq:vi_fcm}
\begin{aligned}
v_{i} = \frac{\sum_{k=1}^{n} u_{ik}^{m} x_{k}}{\sum_{k=1}^{n} u_{ik}^{m}}, \quad \forall i = 1, \ldots, c
\end{aligned}
\end{equation}

In this way, FCM aims to find optimal cluster centers by minimizing intra-cluster distances and maximizing inter-cluster distances using membership values for unlabeled data. 

With the rise of data labeling methods like crowdsourcing \cite{zhang2016learning}, semi-automated \cite{desmond2021semi}, and automated labeling \cite{zhang2021survey}, partially labeled datasets are common, presenting an opportunity for semi-supervised clustering.

\subsection{SSFCM}

To use the full power of prior knowledge, some distinct semi-supervised clustering methods have been proposed \cite{bair2013semi}. In 1997 researchers proposed SSFCM, a method that computes a fidelity term considering the partially labeled instances of a dataset X \cite{pedrycz1997fuzzy}. Given that now X comprises labeled and unlabeled data, being the first $l$ instances of X labeled with $y_k \big|_{k=1}^{l} \in \{1, \ldots, c\}$, and the $(n -l)$ remaining unlabeled. Equation \eqref{eq:ss} represents how the fidelity term is computed: 

\begin{equation}\label{eq:ss}
\mathcal{R} = \sum_{k=1}^{n} \sum_{i=1}^{c} (u_{ik} - f_{ik} b_{k})^m d_{ik}^2
\end{equation}

Here, B is an array of size $n$ indicating if an instance is labeled or not. B $= [b_k]_{n}$, $b_k = 1$ if the $k^{\text{th}}$ instance is labeled and $b_k = 0$, otherwise. $F = [f_{ik}]_{c \times n}$ represents the membership degrees of the labeled instances, with $f_{ik} = 1$ if $i = y_k$ and $f_{ik} = 0$ otherwise. Therefore, this expression computes how different a data point membership is from its indicated label, in the case of a labeled instance.

Then, the objective function is modified to include this characteristic:

\begin{equation}\label{eq:ssfcm}
\begin{aligned}
& J_{SSFCM}(V, U) = J_{FCM}(V, U) + \alpha \mathcal{R}  \quad \\
\end{aligned}
\end{equation}

The parameter $\alpha$ controls the relative importance of the fidelity term $\mathcal{R}$ in the objective function. On the one hand, a larger value of $\alpha$ increases the influence of the fidelity term, making the algorithm prioritize fitting the labeled data points more closely to their indicated labels. On the other hand, a smaller value of $\alpha$ reduces the influence of the fidelity term, making the algorithm rely more on the unsupervised clustering objective.

Consequently, having clustering algorithms trusting and prioritizing labeled data on the objective function and the emergence of so many forms to label data brings forth a critical question: how trustworthy are these provided labels?

With labels potentially being generated through different processes, the accuracy and reliability of the labels can vary significantly, leading to a heightened risk of incorporating mislabeled or noisy data into the algorithm. This uncertainty raises important concerns about the degree to which clustering algorithms should depend on these labels, especially when the integrity of the data cannot always be guaranteed.

\subsection{AS3FCM}

To mitigate the impact of wrong labels on clustering accuracy researchers have developed the concept of safety degrees \cite{GAN2019107}. The many variations of safety semi-supervised clustering algorithms that have appeared ultimately \cite{bair2013semi} have also inspired this paper — in particular, the work developed by H.Gan, Z.Yang, and R.Zhou in 2023 called AS3FCM \cite{gan2023adaptive}, served as the basis for this research. 

In AS3FCM researchers proposed an improved S3C algorithm capable of adaptatively estimating the safety degrees correlating it with the differences of the labeled instances and their $p$ nearest unlabeled neighbors. Firstly, a local graph is built using a fixed number of $p$UN for each labeled instance. 

\begin{equation}\label{eq::wkr}
w_{kr} = \begin{cases} 
\exp \{ - \frac{\| x_k - x_r \|_2^2}{\sigma^2} \}; & \text{if } x_r \in N_p(x_k) \\
0; & \text{otherwise}
\end{cases}
\end{equation}

Here, $x_k$ represents the labeled instance $k$, $x_r$ is an unlabeled instance, and $N_p(x_k)$ indicates a subset of the $p$ unlabeled neighbors of $x_k$. The distances between labeled and unlabeled instances are computed using the Euclidean method, and $\sigma$ is set to the average distance between the instances.

Also, a local consistency is added to the SSFCM objective function \eqref{eq:ssfcm}. To safely handle mislabeled instances, the authors use local consistency to estimate safety degrees.

In   \eqref{eq::as3fcm} $s_k$ denotes the safety degree, $\lambda_1$ and $\lambda_2$ are regularization parameters that weigh the differences between the membership of labeled instances and their indicated label, and the differences of labeled instances and their unlabeled neighbors respectively. Notice that the summation of $s_k$ is constrained to 1.

\begin{equation}\label{eq::as3fcm}
\begin{aligned}
\min_{u_{ik}, v_i, s_k} \; J_a = & \sum_{k=1}^{n} \sum_{i=1}^{c} u_{ik}^m d_{ik}^2 + \lambda_1 \sum_{k=1}^{l} s_k \sum_{i=1}^{c} (u_{ik} - f_{ik})^2 d_{ik}^2 \\
& + \lambda_2 \sum_{k=1}^{l} \left( s_k \sum_{r=l+1}^{n} w_{kr} \sum_{i=1}^{c} (f_{ik} - u_{ir})^2 \right) \\
& + \left( \frac{2}{s_k + 1} - 1 \right) \sum_{r=l+1}^{n} w_{kr} \sum_{i=1}^{c} (u_{ik} - u_{ir})^2 \\
\text{s.t.} \; & \sum_{i=1}^{c} u_{ik} = 1, \; \forall k = 1, \ldots, n \\
& \sum_{k=1}^{l} s_k = 1 \\
& 0 \leq u_{ik} \leq 1, \; \forall i = 1, \ldots, c, \; k = 1, \ldots, n \\
& 0 \leq s_k \leq 1, \; \forall k = 1, \ldots, l
\end{aligned}
\end{equation}

Then, the parameters $u$, $v$, and $s$ are iteratively computed by expressions resulting from the solved derivatives of relevant parts of $J_a$ with respect to each one. Notice that the membership is calculated differently for labeled $u_{ik}$ and unlabeled $u_{ir}$. The full details on how to get to them are described in \cite{gan2023adaptive} and it is beyond the scope of this paper to dive deeper. 

Updating labeled instances $u_{ik}$: 
\begin{equation}\label{eq::as3uik}
u_{ik} = \frac{p_{ik} + \frac{1 - \sum_{i=1}^{c} \frac{p_{ik}}{q_{ik}}}{\sum_{i=1}^{c} \frac{1}{q_{ik}}}}{q_{ik}}
\end{equation}
where $p_{ik} = \lambda_1 s_k f_{ik} d_{ik}^2 + \lambda_2 \left( \frac{2}{s_k + 1} - 1 \right) \sum_{r=l+1}^{n} w_{kr} u_{ir}, \quad$ and $q_{ik} = d_{ik}^2 + \lambda_1 s_k d_{ik}^2 + \lambda_2 \left( \frac{2}{s_k + 1} - 1 \right) \sum_{r=l+1}^{n} w_{kr}$.

Updating unlabeled instances $u_{ir}$: 
\begin{equation}\label{eq::as3uir}
u_{ir} = \frac{z_{ir} + \frac{1 - \sum_{i=1}^{c} \frac{z_{ir}}{t_{ir}}}{\sum_{i=1}^{c} \frac{1}{t_{ir}}}}{t_{ir}}
\end{equation}

where $z_{ir} = \lambda_2 \sum_{k=1}^{l} \left( s_k w_{kr} f_{ik} + \left( \frac{2}{s_k + 1} - 1 \right) w_{kr} u_{ik} \right)$, and  $t_{ir} = d_{ir}^2 + \lambda_2 \sum_{k=1}^{l} \left( s_k w_{kr} + \left( \frac{2}{s_k + 1} - 1 \right) w_{kr} \right).$

Updating cluster center $v_i$: 
\begin{equation}\label{eq::vi}
\begin{aligned}
v_i &= \frac{\sum_{k=1}^{n} u_{ik}^2 x_k + \lambda_1 \sum_{k=1}^{l} s_k (u_{ik} - f_{ik})^2 x_k}{\sum_{k=1}^{n} u_{ik}^2 + \lambda_1 \sum_{k=1}^{l} s_k (u_{ik} - f_{ik})^2}
\end{aligned}
\end{equation}

Finally, updating $s_k$: 

\begin{equation}\label{eq:20}
\begin{aligned}
J_d = \frac{1}{2} \sum_{k=1}^{l} \Omega_k s_k^2 + \sum_{k=1}^{l} \Delta_k s_k \\
\end{aligned}
\end{equation}

\begin{equation*}
\begin{aligned}
\quad \text{s.t.}  \sum_{k=1}^{l} s_k = 1
\end{aligned}
\end{equation*}

\begin{equation*}
0 \leq s_k \leq 1, \quad \forall k = 1, \ldots, l
\end{equation*}

Here,
\begin{equation*}
\Omega_k = 4\lambda_2 \sum_{r=l+1}^{n} w_{kr} \sum_{i=1}^{c} (u_{ik} - u_{ir})^2
\end{equation*}

and
\begin{equation*}
\begin{aligned}
\Delta_k &= \lambda_1 \sum_{i=1}^{c} (u_{ik} - f_{ik})^2 d_{ik}^2 \\
&\quad + \lambda_2 \sum_{r=l+1}^{n} w_{kr} \sum_{i=1}^{c} (f_{ik} - u_{ir})^2 \\
&\quad - 2\lambda_2 \sum_{r=l+1}^{n} w_{kr} \sum_{i=1}^{c} (u_{ik} - u_{ir})^2.
\end{aligned}
\end{equation*}

\section{KNN Graph-Based Safe SSFCM (K-GBS3FCM)}

This section describes the proposed modifications made on AS3FCM resulting in the K-GBS3FCM algorithm.

\subsection{Dynamic number of unlabeled neighbors}

The first step in the method is to construct a graph $W$ in order to estimate the local consistency for each labeled instance. This is done as shown in   \eqref{eq::wkr}.

It is important to note that different datasets may exhibit unique and complex local structures, which can be difficult to capture using a fixed number of unlabeled neighbors to assess the quality of the safety degree. For example, a fixed $p$UN might include connections with instances from other clusters in regions with high density, negatively affecting the safety degree. Conversely, in sparse regions, it might not capture enough connections to properly evaluate the local structure, resulting in an unreliable safety degree. Additionally, a fixed $p$UN can include outliers in the neighborhood that are not representative of the cluster, distorting the safety degree and jeopardizing clustering accuracy. 

Different from AS3FCM, this work uses the KNN method to estimate a dynamic $p$UN for each labeled instance $x_k$, denoted as $dp$UN$_k$. This is done using a proxy for density estimation. 

So, let UN be the set of unlabeled instances, and let $d(x_k, xu_i)$ be the distance between the labeled instance $x_k$ and the unlabeled instance $xu_i \in$ UN. Here, KNN is used to compute the set of the K nearest unlabeled neighbors for each labeled data point. The average distance \( \bar{d}_{Kk} \) of the labeled instance \( x_k \) to its \( K \) nearest unlabeled neighbors is given by:

\begin{equation}
\begin{aligned}
\text{KNN}(K, x_k) &= \{xu_1, xu_2, \ldots, xu_K\} \\
\bar{d}_{Kk} &= \frac{1}{K} \sum_{i=1}^{K} d(x_k, xu_{i})
\end{aligned}
\end{equation}

Here, $\bar{d}_K$ serves as a proxy for density, meaning that regions with high density will have a small $\bar{d}_K$ and regions with low density will have a high $\bar{d}_K$. Therefore, the set $\bar{D} = \{\bar{d}_{K1}, \bar{d}_{K2},...,\bar{d}_{Kl} \}$ of size $l$, as the number of labeled instances, is computed.

However, to improve the algorithm's performance, the authors limit $dp$UN to a minimum and a maximum number of neighbors, denoted as UN$_{min}$ and UN$_{max}$, respectively.

\begin{equation}\label{eq::dpunj}
\begin{aligned}
\small
dp{\text{UN}}_k &= \text{UN}_{\min} + (\text{UN}_{\max} - \text{UN}_{\min}) \times \frac{\bar{d}_{Kk} - \min(\bar{D})}{\max(\bar{D}) - \min(\bar{D})}
\end{aligned}
\end{equation}

This way, $dp{\text{UN}}_k$ is computed for each $x_k$ in labeled instances. This dynamic number of unlabeled neighbors is also used to compute the safety degree.

\subsection{Safety degree estimation}

For this work, the authors have decided to experiment with a simplified version of local consistency estimation. Having $s_k$ as part of the optimization function results in a complex solution since the optimization problem (OP) for it is non-convex, making it difficult to solve with a gradient approach. In AS3FCM the authors utilized Taylor expansion and quadratic programming to address this issue, whereas in K-GBS3FCM the authors wanted to investigate whether a simpler approach could also yield good results. 

Initially, KNN is once again used to retrieve the $dp$UN nearest unlabeled neighbor instances for each labeled one. 
After that, the local inconsistency $li$ for each labeled instance is measured as follows: 

\begin{equation}
\begin{aligned}
\text{KNN}(dp{\text{UN}}_k, x_k) &= \{xu_1, xu_2, \ldots, xu_{dp{\text{UN}}_k} \} \\
li_k &= \sum_{r=1}^{dp{\text{UN}}_k} \sum_{i=1}^{c}  \| u_{ir} - f_{ik} \|
\end{aligned}
\end{equation}

Where $Li = [li_{k}]_{1\times l}$, and $li_k$ defines the local inconsistency for labeled instance $x_k$. Subsequently, the local consistency is computed as shown in   \eqref{eq::sj}.

\begin{equation}\label{eq::sj}
\begin{aligned}
s_k &= \frac{1}{1 + \frac{li_k}{dp\text{UN}_k }} 
\end{aligned}
\end{equation}

As can be seen in   \eqref{eq::sj} the inverse relationship with $li_k$ means that higher inconsistency leads to a lower score $s_k$ and lower inconsistency leads to a higher score $s_k$. Therefore high consistency scores might indicate that the labeled instance fits well within the cluster and its unlabeled neighbors and low consistency might indicate an outlier, misclassified, or even mislabeled instance. 

\subsection{Increased influence of safety degree}

Primarily, it is paramount to understand the role of the regularization parameters $\lambda1$ and $\lambda2$ presented in   \eqref{eq::as3fcm}. On the one hand, $\lambda1$ is strongly associated with labeled data in the objective function. This means that higher $\lambda1$ emphasizes the difference between membership values of labeled data and their expected membership value concerning the expected label. Consequently, the algorithm strives to closely match the labeled data to their expected values.

On the other hand, $\lambda2$ relates more to the relationship between labeled and unlabeled data, meaning that increasing the power of $\lambda2$ will make the algorithm emphasize the minimization of the difference between membership values of unlabeled data and labeled data, altogether. Therefore this would naturally try to approximate the membership values of labeled, unlabeled, and expected labeled data points. 

The authors aimed to investigate the relationship between local consistency and unlabeled data. They assumed that when the safety degree is reliable, the algorithm should emphasize the part of the objective function in   \eqref{eq::as3fcm} that computes the difference between unlabeled and labeled data, specifically the part multiplied by $\lambda2$. Increasing this part of the equation directs the objective function to minimize the difference between unlabeled and labeled data. This approach is logical because, with a consistent degree of safety for a given labeled data point, it is reasonable to approximate the membership values of its nearest unlabeled neighbors to its own membership values.

First, a step function is executed on the average safety degrees.

\begin{equation}\label{eq::step}
Step(\bar{s}) = \begin{cases} 
1; & \text{if } \bar{s} \geq \theta_s \\
0; & \text{otherwise}
\end{cases}
\end{equation}

Here,  $\bar{s}$ is the mean of the safety degrees of the labeled instances, and $\theta_s$ represents the safety degree threshold. Next, this value is used to activate a multiplier of $\lambda2$ part of   \eqref{eq::as3fcm}.

\begin{equation}\label{eq::kgbs3fcm}
\begin{aligned}
\min_{u_{ik}, v_i} \; J_{a'} = & \sum_{k=1}^{n} \sum_{i=1}^{c} u_{ik}^m d_{ik}^2 + \lambda_1 \sum_{k=1}^{l} s_k \sum_{i=1}^{c} (u_{ik} - f_{ik})^2 d_{ik}^2 \\
& + \lambda_2 \lambda_1^{Step(\bar{s})} \sum_{k=1}^{l} \left( s_k \sum_{r=l+1}^{n} w_{kr} \sum_{i=1}^{c} (f_{ik} - u_{ir})^2 \right) \\
& + \left( \frac{2}{s_k + 1} - 1 \right) \sum_{r=l+1}^{n} w_{kr} \sum_{i=1}^{c} (u_{ik} - u_{ir})^2 \\
\text{s.t.} \; & \sum_{i=1}^{c} u_{ik} = 1, \; \forall k = 1, \ldots, n \\
& \sum_{k=1}^{l} s_k \leq l \\
& 0 \leq u_{ik} \leq 1, \; \forall i = 1, \ldots, c, \; k = 1, \ldots, n \\
& 0 \leq s_k \leq 1, \; \forall k = 1, \ldots, l
\end{aligned}
\end{equation}

\subsection{K-GBS3FCM Solution}
The solution combines everything in an attempt to improve clustering accuracy. Unlike AS3FCM, local consistency minimization is not included as part of the objective function. 

\begin{algorithm}
\caption{K-GBS3FCM}
\label{algo::kgb}
\begin{algorithmic}[1]
\State \textbf{Input:} A labeled subset $X_l = [x_1, \cdots, x_l]$ with the corresponding labels $Y_l = [y_1, \cdots, y_l]^T$ and an unlabeled subset $X_u = [x_{l+1}, \cdots, x_n]$. The parameters $\lambda_1, \lambda_2, K, \theta_s, \text{UN}_{min}, \text{UN}_{max}, \eta$, and $Maxiter$.
\State \textbf{Output:} Optimal partition matrix $U^*$.
\State Get $dp$UN$_k$ for each labeled instance $x_k$ using   \eqref{eq::dpunj};
\State Construct the graph $W$ with   \eqref{eq::wkr};
\State Initialize the cluster center $V^{(0)}$ by computing the mean of the labeled samples in each class;
\State Initialize $U^{(0)}$, labeled and unlabeled;
\For {$t = 1 : Maxiter$}
    \State Update $u_{ik}^{(t)}$ and $u_{ir}^{(t)}$ using   \eqref{eq::kgbs3uij} and   \eqref{eq::kgbs3uir};
    \State Update $v_i^{(t)}$ using   \eqref{eq::vi};
    \State Update $s_k^{(t)}$ using   \eqref{eq::sj};
    \State Compute $\bar{s}$;
    \State Compute the value of $J_{a'}^{(t)}$ with   \eqref{eq::kgbs3fcm};
    \If {$|J_{a'}^{(t)} - J_{a'}^{(t-1)}| < \eta$}
        \State \Return $U^*$;
    \EndIf
\EndFor
\State \Return $U^*$;
\end{algorithmic}
\end{algorithm}

An analysis of complexity for the algorithm \eqref{algo::kgb} can be found in \cite{gan2023adaptive}.
Notice that in this work the summation of all safety degrees is not constrained to 1. The authors believe that constraining the total sum of safety degrees to 1 might not allow sufficient differentiation between labeled instances of varying reliability. This could restrain flexibility because instances that should have high values of safety degree may end up with lower values due to tight normalization. 

As shown in \eqref{eq::kgbs3fcm} if the average local consistency is sufficiently reliable, the $\lambda2$ part of the objective function is amplified by $\lambda1$. The algorithm then iteratively runs in order to minimize variables  $u_{ik}$ and $v_i$.

\subsubsection{Updating membership}

Changing the objective function reflects on its derivative and, consequently, on how membership matrix U is updated. As a result, the following equations, which take into account the increased influence of safety degrees, are obtained from   \eqref{eq::as3uik} and \eqref{eq::as3uir}, respectively:

Updating labeled instances $u_{ik}$: 
\begin{equation}\label{eq::kgbs3uij}
\begin{aligned}
u_{ik} &= \frac{p_{ik} + \frac{1 - \sum_{i=1}^{c} \frac{p_{ik}}{q_{ik}}}{\sum_{i=1}^{c} \frac{1}{q_{ik}}}}{q_{ik}}
\end{aligned}
\end{equation}

where 
\begin{align*}
p_{ik} &= \lambda_1 s_k f_{ik} d_{ik}^2 + \lambda_2 \lambda_1^{\text{Step}(\bar{s})}  \left( \frac{2}{s_k + 1} - 1 \right) \sum_{r=l+1}^{n} w_{kr} u_{ir} \quad \\
q_{ik} &= d_{ik}^2 + \lambda_1 s_k d_{ik}^2 + \lambda_2 \lambda_1^{\text{Step}(\bar{s})} \left( \frac{2}{s_k + 1} - 1 \right) \sum_{r=l+1}^{n} w_{kr}
\end{align*}

Updating unlabeled instances $u_{ir}$: 
\begin{equation}\label{eq::kgbs3uir}
\begin{aligned}
u_{ir} &= \frac{z_{ir} + \frac{1 - \sum_{i=1}^{c} \frac{z_{ir}}{t_{ir}}}{\sum_{i=1}^{c} \frac{1}{t_{ir}}}}{t_{ir}}
\end{aligned}
\end{equation}

where
\begin{align*}
z_{ir} &= \lambda_2 \lambda_1^{\text{Step}(\bar{s})}  \sum_{k=1}^{l} \left( s_k w_{kr} f_{ik} + \left( \frac{2}{s_k + 1} - 1 \right) w_{kr} u_{ik} \right) \\
t_{ir} &= d_{ir}^2 + \lambda_2 \lambda_1^{\text{Step}(\bar{s})} \sum_{k=1}^{l} \left( s_k w_{kr} + \left( \frac{2}{s_k + 1} - 1 \right) w_{kr} \right)
\end{align*}

The centers of the clusters are computed the same as previously shown in   \eqref{eq::vi}

It is crucial to observe that $U^{(0)}$ is initialized differently for labeled and unlabeled instances. For labeled ones, it sets the value 1 to the class given by the inputted labels $Y_l$. This seems reasonable once the process is semi-supervised and the labels are being inputted it means that these data points should belong to their corresponding labels. For the unlabeled ones, memberships are initialized using the traditional FCM method. 

The parameter $K$ here represents the initial number of neighbors the algorithm K-Means should consider to compute the average distance and, thus, estimate density. 

\subsection{Complexity Analysis}
The time complexity of the K-GBS3FCM algorithm is driven by several key operations: the K-Nearest Neighbors (KNN) search, membership updates from the Fuzzy C-Means (FCM) method, safety degree estimation, and cluster center updates. The KNN search, used to dynamically estimate neighbors for each labeled data point, has a complexity of 
$O(l(n-l)d)$, where $l$ is the number of labeled points, $n$ is the total number of data points, and $d$ is the data dimensionality. Updating the membership values for both labeled and unlabeled data takes 
$O(ncd)$, where $c$ is the number of clusters. Safety degree estimation adds 
$O(lcp_{UN})$, where $p_{UN}$ is the number of nearest neighbors considered for each labeled point, and updating the cluster centers takes 
$O(cnd)$. These steps are repeated over $T$ iterations, leading to an overall complexity of 
$O(T(nld+ncd))$. This means that K-GBS3FCM would typically perform better for large datasets with many labeled points or clusters compared to \cite{gan2023adaptive}.

The space complexity of K-GBS3FCM is primarily driven by storing the dataset and membership matrix. Storing $n$ data points of dimensionality $d$ requires $O(nd)$, while the membership matrix, with $n$ points and $c$ clusters, takes $O(nc)$. Additional space is needed for cluster centers $O(cd)$, the KNN graph $O(lp_{UN})$ for labeled points $l$ and neighbors $p_{UN}$, and safety degrees $O(l)$. The total space complexity is $O(nd + nc + cd + lp_{UN})$, which can be written as $O(nd + nc)$, ensuring scalability for large datasets with a moderate number of clusters.

\section{Experiment and Results}

The primary goal of the experiments was to evaluate K-GBS3FCM's performance in achieving competitive clustering accuracy, even with mislabeled data. The authors hypothesize that a dynamic neighborhood, simplified safety degrees, and stronger local consistency on unlabeled data could enhance accuracy and outperform both traditional and new semi-supervised algorithms.

All the code developed and results found during this work are publicly available\footnote{\href{https://github.com/gabrielmsantos/GBS3FCM}{https://github.com/gabrielmsantos/GBS3FCM}}.

\subsection{Datasets}
The datasets used to experiment and assess the performance of the proposed work were exactly the same as those reported in \cite{gan2023adaptive}. A total of eight datasets were utilized, of which three (bupa\cite{misc_liver_disorders_60}, dermatology\cite{misc_dermatology_33}, and waveform\cite{misc_waveform_database_generator_(version_1)_107}) were extracted from the University of California Irvine (UCI) Machine Learning Repository \cite{frank2010uci}, and three others (diabetes \cite{kaggle_diabetes}, heart \cite{misc_heart_disease_45}, and wdbc\cite{kaggle_wdbc}) were extracted from Kaggle datasets repository \cite{kaggle_2024}. 

The gauss50 and gauss50x datasets were both generated using multivariate normal distributions with identical mean vectors for their respective classes (0.25 for class 1 and -0.25 for class 2 across all features). However, the key difference lies in the data generation process. In gauss50, each class is generated from a single normal distribution, resulting in distinct, well-separated clusters. In contrast, gauss50x is created using a mixture of two multivariate normal distributions for each class, with slightly different sample proportions, leading to more complex and overlapping clusters. This distinction in data structure increases the difficulty of separating the classes in gauss50x compared to gauss50, despite their similar means.

It is important to note that Bupa dataset should be used exactly as the owners advise in the instructions, using five features, and the sixth one (drinks) as the target class \cite{turney1994cost}. The classification rule applied to the Bupa dataset is as follows:

\[
\text{Bupa-Class}(x) = 
\begin{cases} 
      0; & \text{if } x_6 < 3 \\
      1; & \text{otherwise}
   \end{cases}
\]
where \( x_6 \) represents the value in the 6th column (drinks) of the dataset. Further details of datasets are shown in table \eqref{tab:datasets}. 

\begin{table}[h!]
    \centering
    \resizebox{\columnwidth}{!}{%
    \begin{tabular}{|c|c|c|c|}
        \hline
        \textbf{Dataset name} & \textbf{Number of data} & \textbf{Number of features} & \textbf{Number of classes} \\
        \hline
        Bupa         & 345  & 5   & 2 \\
        \hline
        Dermatology  & 358  & 33  & 6 \\
        \hline
        Diabetes     & 768  & 8   & 2 \\
        \hline
        Gauss50      & 1550 & 50  & 2 \\
        \hline
        Gauss50x     & 2000 & 50  & 2 \\
        \hline
        Heart        & 297  & 13  & 2 \\
        \hline
        Waveform     & 5000 & 21  & 3 \\
        \hline
        Wdbc         & 568  & 30  & 2 \\
        \hline
    \end{tabular}
    }
    \caption{Details of experimental datasets.}
    \label{tab:datasets}
\end{table}

\subsection{Benchmarking Algorithms }

To evaluate the performance of the proposed algorithm, a comparison is made with eight algorithms reported in the paper AS3F3CM, from H.Gan, Z.Yang and R.Zhou. \cite{gan2023adaptive} that had the best results. The author of paper \cite{gan2023adaptive} compared their algorithm against eleven different algorithms. From these, eight (K-Means\cite{Kmeans}, FCM\cite{FCM}, SMKFCM\cite{SMKFCM}, SMKFC-ER\cite{SALEHI2021667}, CSNMF\cite{PENG2021107683}, LHC-S$^3$-FCM\cite{GAN2018384}, CS3FCM\cite{GAN2019107}, and AS3FCM\cite{gan2023adaptive}) have been selected for use as benchmarks in this study.

The results of these five algorithms, as reported in \cite{gan2023adaptive}, provide a robust benchmark for assessing the effectiveness of the proposed approach. These algorithms have not been rerun; instead, the results published in \cite{gan2023adaptive} are relied upon for the comparison. 

\subsection{Experimental Setup}

To maintain comparability of achievement with other works \cite{gan2023adaptive} \cite{gan2019safe}, the datasets are randomly divided into $20\% $ of labeled data, and  $80\% $ of unlabeled data.
To analyze the impact of incorrect labels, incorrect labels are intentionally assigned to certain instances from the labeled subset, making their assigned labels differ from their ground truth. The proportion of these mislabeled instances ranges from 0\% to 30\%, increasing in 5\% increments. 

As for the parameters for K-GBS3FCM, $\lambda1$ and $\lambda2$ were searched in a combinatorial way within the interval $\{10^{-3}, 10^{-2}, 10^{-1}, 1, 10, 10^{2}\}$. The initial parameter $K$, which defines the neighborhood, was set to 5 for computing the average distance and subsequently the density. The threshold $\theta_s$ was set to 0.6, UN$_{min}$ and UN$_{max}$ assumed the values of 5 and $\sqrt{n}$, respectively. The parameter $\eta$ was set $10^{-4}$, and finally, $Maxiter$ was set to 100.  
 
This procedure was repeated 20 times for each mislabeling configuration to ensure robustness. The tests were conducted on a Mac M3 Max with 48GB of memory and 16 cores. The average execution time for 20 runs across 36 combinations of lambda configurations and all seven mislabeling values varied depending on the dataset. Overall, it took approximately one week to complete the tests on the eight datasets.

It is worth noting that the performance of the algorithms presented here was measured through clustering accuracy. Since K-GBS3FCM is a semi-supervised algorithm, the following equation has been used to compute clustering accuracy (CA).

\begin{equation}\label{eq::ca}
\begin{aligned}
\text{CA} = \frac{\sum_{k=1}^{n} \delta (y_k, \tilde{y}_k)}{n}
\end{aligned}
\end{equation}

In this formula, \( y_k \) represents the true label of \( x_k \), while \( \tilde{y}_k \) denotes the predicted label. The function \( \delta(a, b) \) is defined as 1 when \( a = b \), and 0 otherwise.

\subsection{Results per Dataset}
\label{RpD}

\begin{figure*}[htbp]
    \centering
    \begin{subfigure}[b]{0.46\textwidth}
        \centering
        \includegraphics[width=\textwidth]{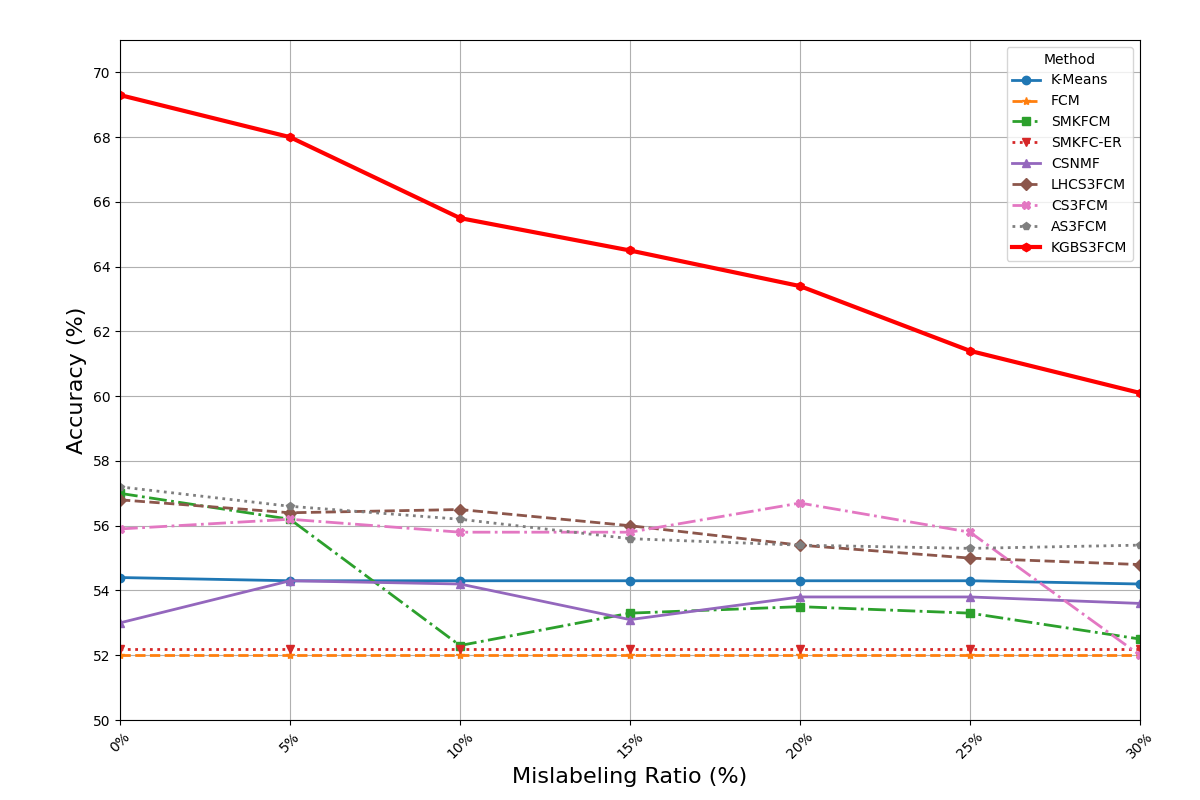} 
        \caption{Bupa}
        \label{fig:bupa}
    \end{subfigure}
    \hfill
    \begin{subfigure}[b]{0.46\textwidth}
        \centering
        \includegraphics[width=\textwidth]{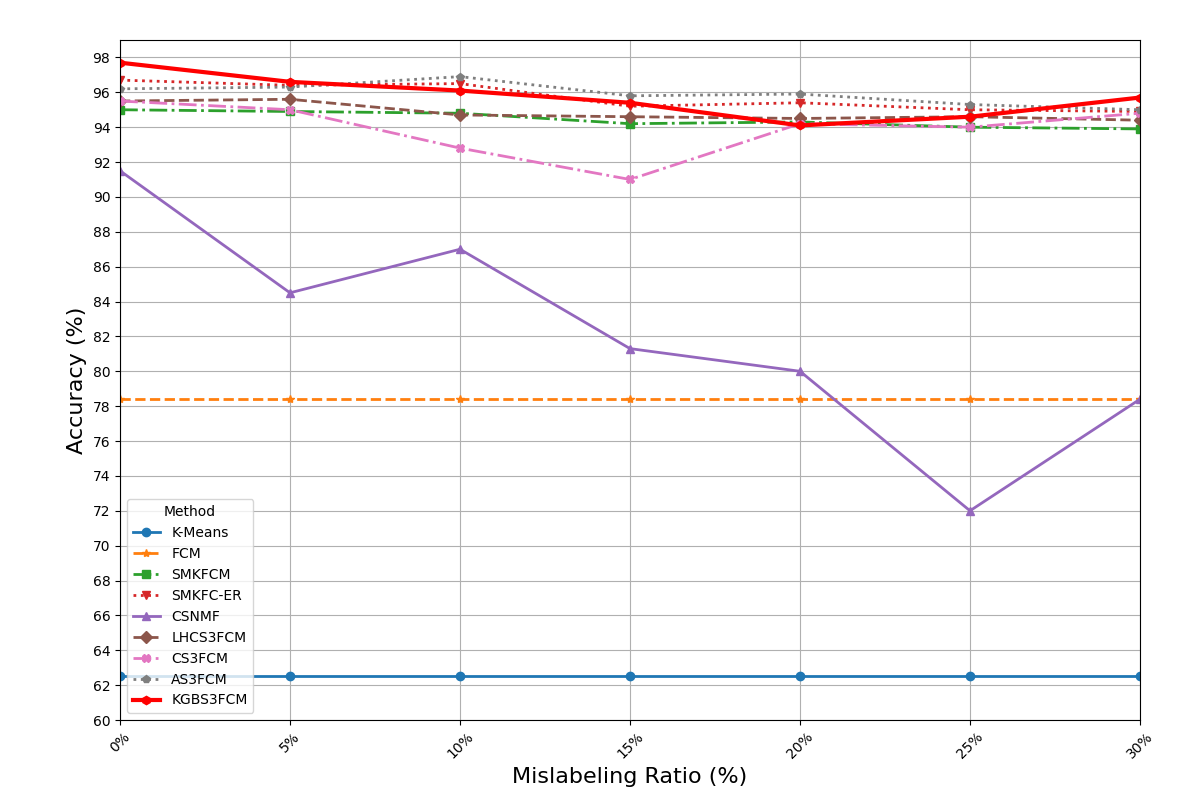} 
        \caption{Dermatology}
        \label{fig:dermatology}
    \end{subfigure}
    \hfill
    \begin{subfigure}[b]{0.46\textwidth}
        \centering
        \includegraphics[width=\textwidth]{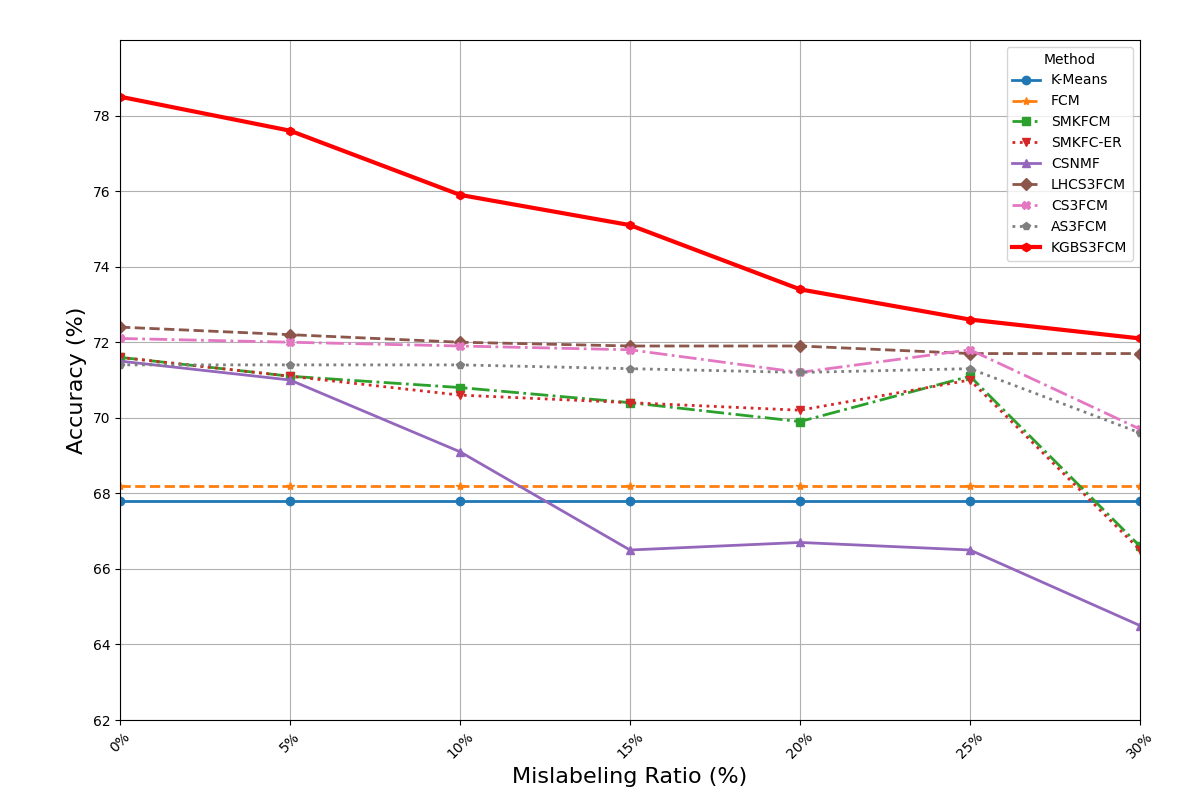} 
        \caption{Diabetes}
        \label{fig:diabetes}
    \end{subfigure}
    \hfill
    \begin{subfigure}[b]{0.46\textwidth}
        \centering
        \includegraphics[width=\textwidth]{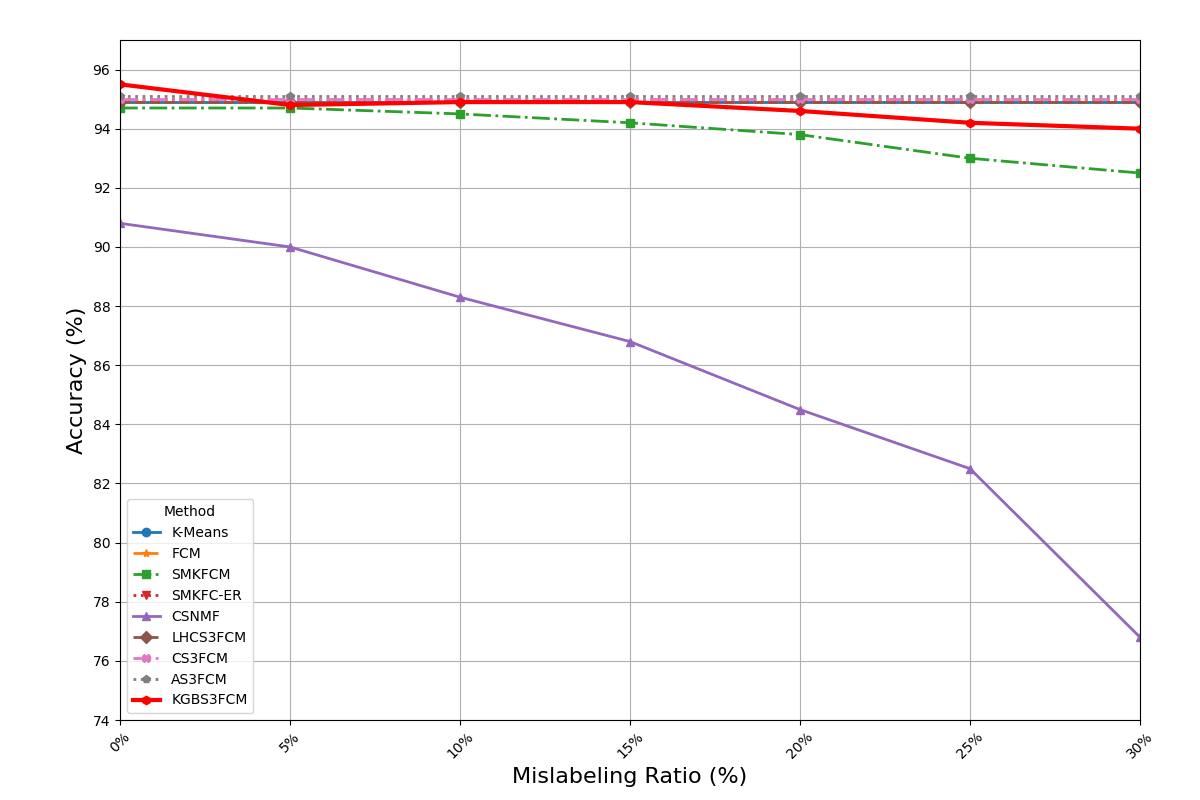} 
        \caption{Gauss50}
        \label{fig:gauss50}
    \end{subfigure}
    \hfill
    \begin{subfigure}[b]{0.46\textwidth}
        \centering
        \includegraphics[width=\textwidth]{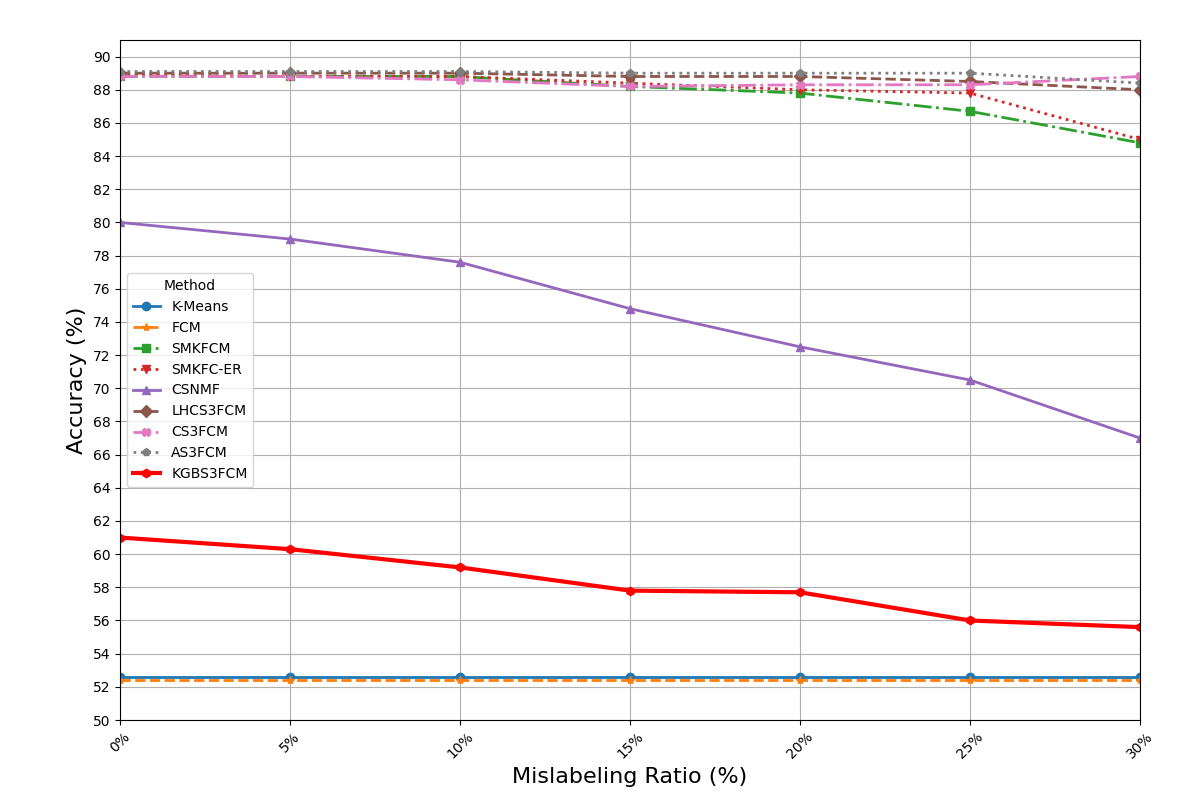} 
        \caption{Gauss50x}
        \label{fig:gauss50x}
    \end{subfigure}
    \hfill
    \begin{subfigure}[b]{0.46\textwidth}
        \centering
        \includegraphics[width=\textwidth]{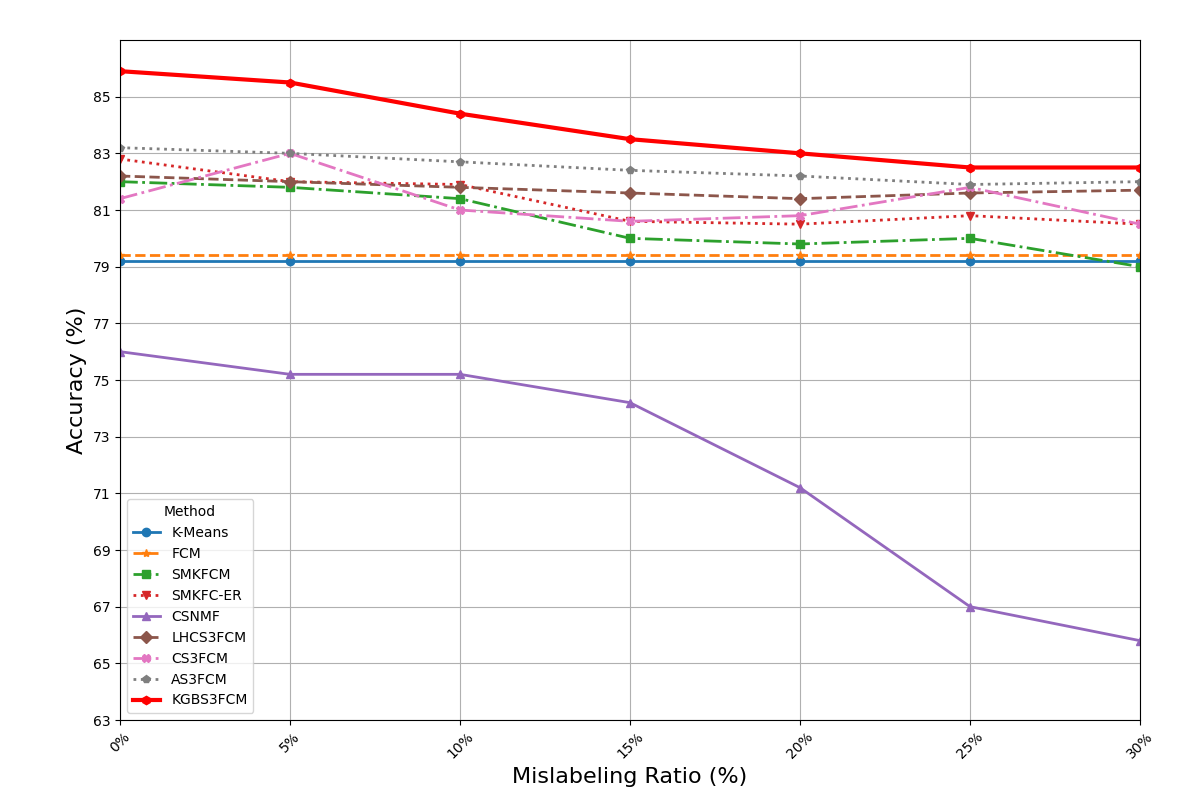} 
        \caption{Heart}
        \label{fig:heart}
    \end{subfigure}
    \hfill
    \begin{subfigure}[b]{0.46\textwidth}
        \centering
        \includegraphics[width=\textwidth]{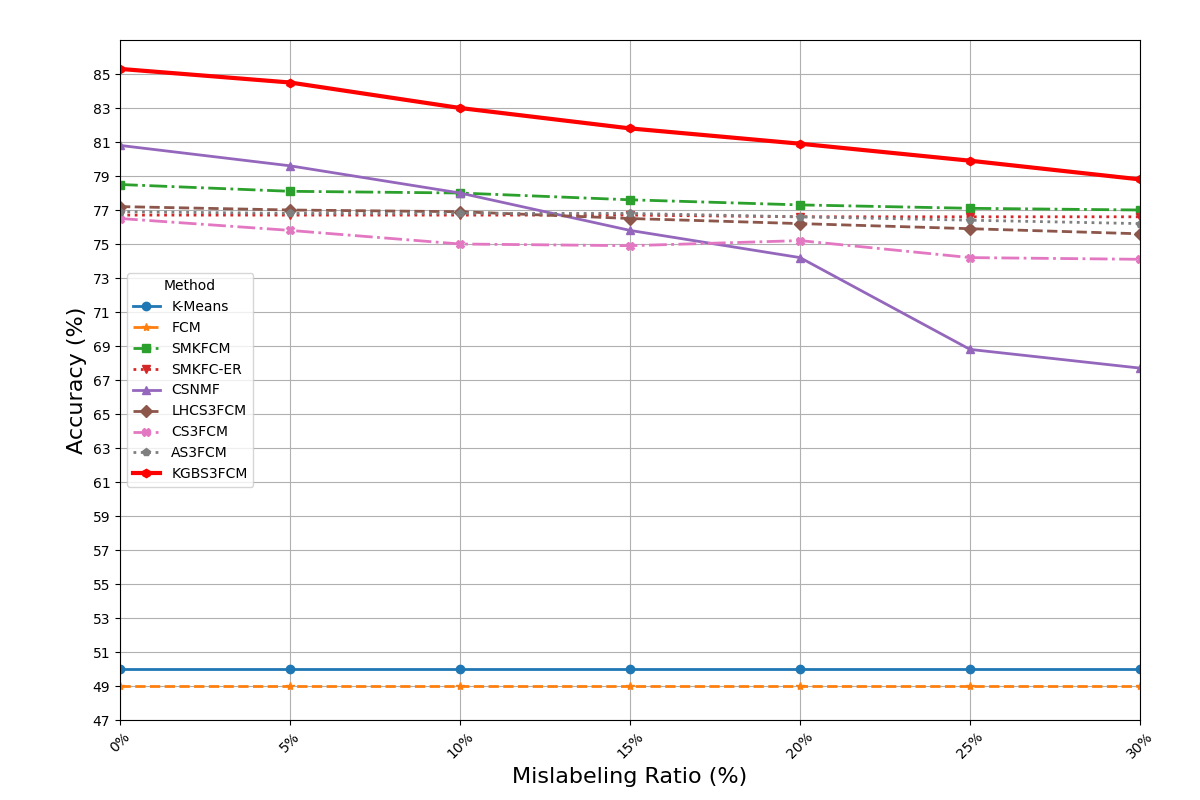} 
        \caption{Waveform}
        \label{fig:waveform}
    \end{subfigure}
    \hfill
    \begin{subfigure}[b]{0.46\textwidth}
        \centering
        \includegraphics[width=\textwidth]{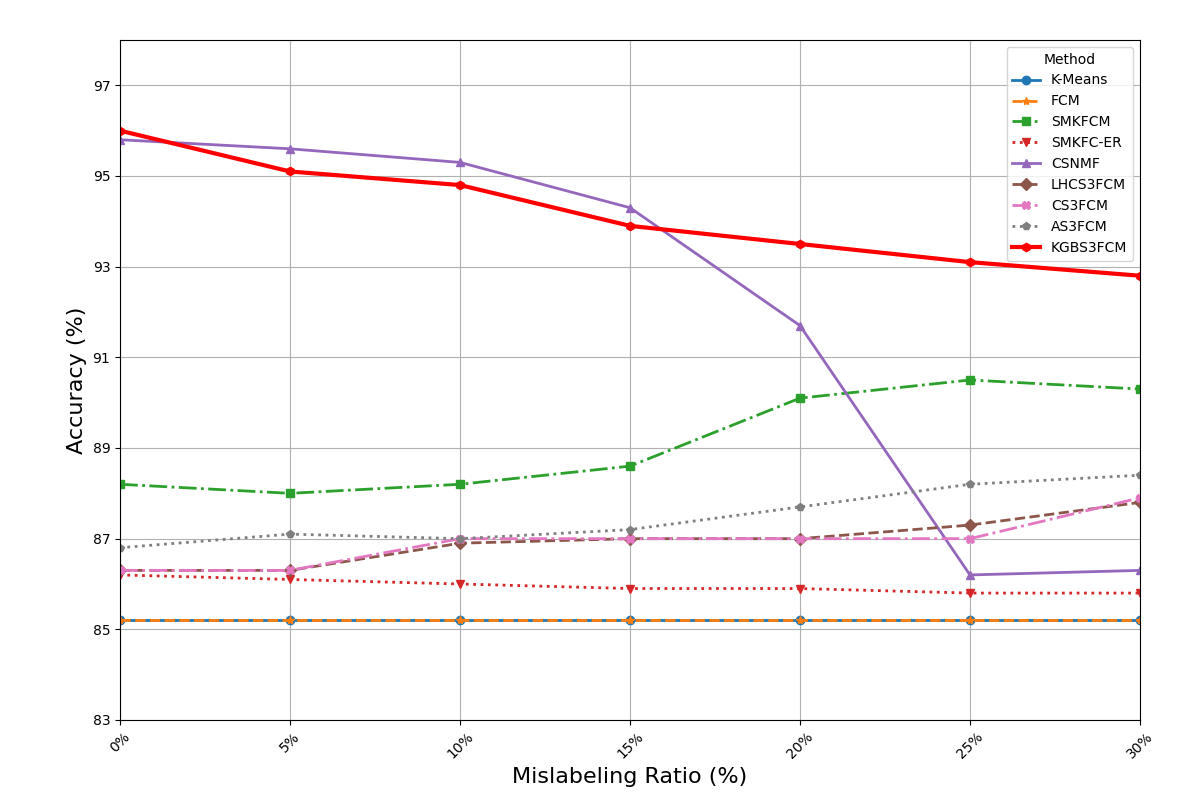} 
        \caption{Wdbc}
        \label{fig:wdbc}
    \end{subfigure}
    \caption{Comparison of algorithms across different datasets}
    \label{fig:comparison}
\end{figure*}

In order to provide a visual aid, results are shown in Fig. \eqref{fig:comparison} which provides a comprehensive comparison of how each algorithm handles increasing amounts of mislabeled data. The specific observations for each dataset and the key findings from these experiments are discussed below.

\subsubsection{Bupa}
Despite strongly decreasing from 69.3\% for no mislabeling to 60.1\% at 30\% of the data is mislabeled, K-GBS3FCM consistently outperforms all other algorithms across all levels of mislabeling. Although having a slight increase from 0\% to 30\%, CSNMF remained at a poor level of accuracy. While some algorithms such as AS3FCM, LHCS3FCM, and CS3FCM had a smaller drop in accuracy when increasing mislabeling instances, they remained significantly lower in accuracy compared to K-GBS3FCM. 

\subsubsection{Dermatology}
Unlike Bupa, K-GBS3FCM demonstrated much more stability in this database as the number of mislabeled instances increased. Additionally, it outperformed the other algorithms in three of the seven categories at 0\%, 5\%, and 30\% mislabeling. Even when it did not achieve the best performance in the remaining categories, K-GBS3FCM showed very close performance and stability, making it a competitive and solid alternative to the other algorithms. 

\subsubsection{Diabetes}
As for the Diabetes database, K-GBS3FCM performed superiorly across all mislabeling levels. However, it suffered from a decrease in clustering accuracy with an increase in mislabeling, going from 78.5\% to 72.1\%. 

\subsubsection{Gauss50}
Here, K-GBS3FCM demonstrated consistency throughout its performance. It begins outperforming the other algorithms with 95.5\% of accuracy at level 0\% then it starts to drop slightly, getting to 94\% for 30\% of mislabeling. Even though, it remains at a competitive level of performance compared to the other semi-supervised algorithms.

\subsubsection{Gauss50x}
Different from previous datasets, the performance of KGBS3FCM on the Gauss50x dataset was significantly poorer compared to its performance on other datasets. While it outperformed the traditional unsupervised methods, K-Means, and FCM, it had the lowest accuracy among the semi-supervised algorithms. The authors hypothesized that this poor performance was due to significant overlap between clusters, caused by the proximity of their means and the spread defined by the covariance matrices.

\subsubsection{Heart}
The performance of KGBS3FCM was once again superior compared to the other algorithms. Its accuracy started at 85.9\% and gradually dropped to 82.5\% when 30\% of the instances were mislabeled. AS3FCM has also remained more stable, with a lower level of clustering accuracy compared to the method presented. In contrast, CSNMF performed worse than the unsupervised methods, highlighting its sensitivity to mislabeled data.

\subsubsection{Waveform}
When executed for the Waveform dataset, K-GBS3FCM reached an average level of accuracy of 85.3\% with no mislabeling dropping down to 78.8\% with 30\% of mislabeled. Its level of accuracy remained superior across all levels of mislabeling. The other methods performed in a stable way, as well, but with inferior level of accuracy compared to the method presented in this work. The unsupervised methods performed substantially worse than the semi-supervised ones. 

\subsubsection{Wdbc}
In the Wdbc dataset, the proposed algorithm outperformed all the other ones at a ratio 0\%, 20\%, 25\%, and 30\% of mislabed instances. CSNMF performed very well with a lower mislabeling ratio. However, it drops significantly as the number of mislabeled instances increases.

\subsection{Analysis of Results}
Table \eqref{tab:accKgb} shows the results of clustering accuracy across all datasets for the proposed method in this work. The green cells represent where the algorithm performed the best compared to the other algorithms. :

\begin{table}[h!]
    \centering
    \resizebox{\columnwidth}{!}{%
    \begin{tabular}{|c|c|c|c|c|c|c|c|}
        \hline
        \textbf{} & \textbf{0\%} & \textbf{5\%} & \textbf{10\%} & \textbf{15\%} & \textbf{20\%} & \textbf{25\%} & \textbf{30\%} \\
        \hline
        Bupa & \cellcolor{green!30}69.3  & \cellcolor{green!30}68 & \cellcolor{green!30}65.5 & \cellcolor{green!30}64.5 & \cellcolor{green!30}63.4 & \cellcolor{green!30}61.4 & \cellcolor{green!30}60.1 \\
        \hline
        Dermatology & \cellcolor{green!30}97.7 & \cellcolor{green!30}96.6 & \cellcolor{yellow!30}96.1 & \cellcolor{yellow!30}95.4 & \cellcolor{yellow!30}94.1 & \cellcolor{yellow!30}94.6 & \cellcolor{green!30}95.7 \\
        \hline
        Diabetes & \cellcolor{green!30}78.5 & \cellcolor{green!30}77.6 & \cellcolor{green!30}75.9 & \cellcolor{green!30}75.1 & \cellcolor{green!30}73.4 & \cellcolor{green!30}72.6 & \cellcolor{green!30}72.1 \\
        \hline
        Gauss50 & \cellcolor{green!30}95.5 & \cellcolor{yellow!30}94.8 &  \cellcolor{yellow!30}94.9 &  \cellcolor{yellow!30}94.9 &  \cellcolor{yellow!30}94.6 &  \cellcolor{yellow!30}94.2 &  \cellcolor{yellow!30}94 \\
        \hline
        Gauss50x &  \cellcolor{yellow!30}61 &  \cellcolor{yellow!30}60.3 &  \cellcolor{yellow!30}59.2 &  \cellcolor{yellow!30}57.8 &  \cellcolor{yellow!30}57.7 &  \cellcolor{yellow!30}56 &  \cellcolor{yellow!30}55.6 \\
        \hline
        Heart & \cellcolor{green!30}85.9 & \cellcolor{green!30}85.5 & \cellcolor{green!30}84.4 & \cellcolor{green!30}83.5 & \cellcolor{green!30}83 & \cellcolor{green!30}82.5 & \cellcolor{green!30}82.5 \\
        \hline
        Waveform & \cellcolor{green!30}85.3 & \cellcolor{green!30}84.5 & \cellcolor{green!30}83 & \cellcolor{green!30}81.8 & \cellcolor{green!30}80.9 & \cellcolor{green!30}79.9 & \cellcolor{green!30}78.8 \\
        \hline
        Wdbc & \cellcolor{green!30}96 & \cellcolor{yellow!30}95.1 & \cellcolor{yellow!30}94.8 & \cellcolor{yellow!30}93.9 & \cellcolor{green!30}93.5 & \cellcolor{green!30}93.1 & \cellcolor{green!30}92.8 \\
        \hline
    \end{tabular}
    }
    \caption{Clustering accuracy (in \%) of K-GBS3FCM for each mislabeling configuration across all datasets.}
    \label{tab:accKgb}
\end{table}

In this proposed experimental setup the performance of K-GBS3FCM has shown to be consistently superior compared to the other unsupervised and supervised methods. Some key observations are highlighted. 

\begin{itemize}
    \item Out of the 56 possible combinations of datasets and mislabeling configurations, the proposed method outperformed all other algorithms 36 times, which is 64.28\% of the cases. In the remaining 35.72\% of the cases, K-GBS3FCM showed competitive performance 23.22\% of the time (in 13 configurations) and inferior performance 12.5\% of the time, all of which were in the Gauss50x dataset.
    
    \item At 0\% mislabeling, where all labels are correct, K-GBS3FCM outperforms all other algorithms across all datasets except Gauss50x. However, even in Gauss50x, it performed better than the unsupervised methods, reinforcing how beneficial accurate labels and prior knowledge are for clustering accuracy.
    
    \item At all levels of mislabeling, the proposed method experienced a loss of accuracy across all datasets. The most significant loss was observed with the Bupa dataset, where accuracy dropped from 69.3\% to 60.1\%, a decrease of 9.2\%. While this highlights the negative impact of mislabeling on performance, it also demonstrates K-GBS3FCM's ability to maintain a relative level of stability.
    
    \item In Gauss50x, K-GBS3FCM only outperformed traditional unsupervised methods. K-GBS3FCM relies on fuzzy memberships, KNN-based neighborhood relationships, and safety degrees to make clustering decisions. Therefore the use of two multivariate normal distributions with close means for each class in Gauss50x significantly increases the likelihood of overlapping clusters \cite{bishop2006pattern} \cite{fraley2002model}.
    As a result, when clusters overlap, memberships become ambiguous, KNN neighbors are drawn from multiple clusters, and safety degrees become less reliable. These factors together make it harder for the algorithm to correctly separate the clusters, leading to reduced classification accuracy
    
    \item With 30\% mislabeled instances, K-GBS3FCM showed superior performance in almost all datasets except for Gauss50 and Gauss50x. This underscores the advantage of using prior knowledge to influence unlabeled data, which can enhance clustering accuracy.
\end{itemize}

\section{Conclusion and Future Work}

The proposed solution in this work integrates dynamic estimation of unlabeled neighbors and simplifies the computation of local consistency. This approach emphasizes the relationship between unlabeled and labeled data, particularly when the average safety degree is reliable. The performance of K-GBS3FCM was evaluated across several datasets, showing significant improvements in clustering accuracy even in the presence of mislabeled instances. Our results demonstrate that K-GBS3FCM outperforms traditional and new semi-supervised clustering algorithms, including AS3FCM, especially as the proportion of mislabeled data increases.

The comparative analysis against existing algorithms further validates the robustness and effectiveness of K-GBS3FCM. The consistent accuracy across diverse datasets highlights its potential for broader application in various domains requiring reliable clustering under noisy conditions. 

Future work will address several key challenges. First, we will explore methods to overcome the problem of poor accuracy in the presence of significant overlapping clusters. Additionally, new forms of density will be computed and investigated to determine the ideal number of neighbors more effectively. We will also study the performance of K-GBS3FCM on higher-dimensional datasets and its sensitivity to different values of $K$ to assess its robustness and adaptability. Finally, further optimizations will be considered to enhance the algorithm's scalability and overall performance.

\bibliography{main.bib}

\begin{thebibliography}{10}
\providecommand{\url}[1]{#1}
\csname url@samestyle\endcsname
\providecommand{\newblock}{\relax}
\providecommand{\bibinfo}[2]{#2}
\providecommand{\BIBentrySTDinterwordspacing}{\spaceskip=0pt\relax}
\providecommand{\BIBentryALTinterwordstretchfactor}{4}
\providecommand{\BIBentryALTinterwordspacing}{\spaceskip=\fontdimen2\font plus
\BIBentryALTinterwordstretchfactor\fontdimen3\font minus \fontdimen4\font\relax}
\providecommand{\BIBforeignlanguage}[2]{{%
\expandafter\ifx\csname l@#1\endcsname\relax
\typeout{** WARNING: IEEEtran.bst: No hyphenation pattern has been}%
\typeout{** loaded for the language `#1'. Using the pattern for}%
\typeout{** the default language instead.}%
\else
\language=\csname l@#1\endcsname
\fi
#2}}
\providecommand{\BIBdecl}{\relax}
\BIBdecl

\bibitem{xu2015comprehensive}
D.~Xu and Y.~Tian, ``A comprehensive survey of clustering algorithms,'' \emph{Annals of data science}, vol.~2, pp. 165--193, 2015.

\bibitem{Kmeans}
\BIBentryALTinterwordspacing
J.~A. Hartigan and M.~A. Wong, ``Algorithm as 136: A k-means clustering algorithm,'' \emph{Journal of the Royal Statistical Society. Series C (Applied Statistics)}, vol.~28, no.~1, pp. 100--108, 1979. [Online]. Available: \url{http://www.jstor.org/stable/2346830}
\BIBentrySTDinterwordspacing

\bibitem{FCM}
J.~C. Bezdek, \emph{Pattern recognition with fuzzy objective function algorithms}.\hskip 1em plus 0.5em minus 0.4em\relax Springer Science \& Business Media, 2013.

\bibitem{GMM}
\BIBentryALTinterwordspacing
Z.~Lu, ``An iterative algorithm for entropy regularized likelihood learning on gaussian mixture with automatic model selection,'' \emph{Neurocomputing}, vol.~69, no.~13, pp. 1674--1677, 2006, blind Source Separation and Independent Component Analysis. [Online]. Available: \url{https://www.sciencedirect.com/science/article/pii/S0925231206000294}
\BIBentrySTDinterwordspacing

\bibitem{gan2013using}
H.~Gan, N.~Sang, R.~Huang, X.~Tong, and Z.~Dan, ``Using clustering analysis to improve semi-supervised classification,'' \emph{Neurocomputing}, vol. 101, pp. 290--298, 2013.

\bibitem{bair2013semi}
E.~Bair, ``Semi-supervised clustering methods,'' \emph{Wiley Interdisciplinary Reviews: Computational Statistics}, vol.~5, no.~5, pp. 349--361, 2013.

\bibitem{grira2004unsupervised}
N.~Grira, M.~Crucianu, and N.~Boujemaa, ``Unsupervised and semi-supervised clustering: a brief survey,'' \emph{A review of machine learning techniques for processing multimedia content}, vol.~1, no. 2004, pp. 9--16, 2004.

\bibitem{pedrycz1997fuzzy}
W.~Pedrycz and J.~Waletzky, ``Fuzzy clustering with partial supervision,'' \emph{IEEE Transactions on Systems, Man, and Cybernetics, Part B (Cybernetics)}, vol.~27, no.~5, pp. 787--795, 1997.

\bibitem{Wang2013SafetyAwareSC}
\BIBentryALTinterwordspacing
Y.~Wang and S.~Chen, ``Safety-aware semi-supervised classification,'' \emph{IEEE Transactions on Neural Networks and Learning Systems}, vol.~24, pp. 1763--1772, 2013. [Online]. Available: \url{https://api.semanticscholar.org/CorpusID:14396569}
\BIBentrySTDinterwordspacing

\bibitem{gan2019safe}
H.~Gan, ``Safe semi-supervised fuzzy c-means clustering,'' \emph{IEEE Access}, vol.~7, pp. 95\,659--95\,664, 2019.

\bibitem{GAN2018384}
\BIBentryALTinterwordspacing
H.~Gan, Y.~Fan, Z.~Luo, and Q.~Zhang, ``Local homogeneous consistent safe semi-supervised clustering,'' \emph{Expert Systems with Applications}, vol.~97, pp. 384--393, 2018. [Online]. Available: \url{https://www.sciencedirect.com/science/article/pii/S0957417417308680}
\BIBentrySTDinterwordspacing

\bibitem{SMKFCM}
\BIBentryALTinterwordspacing
S.~D. Mai and L.~T. Ngo, ``Multiple kernel approach to semi-supervised fuzzy clustering algorithm for land-cover classification,'' \emph{Engineering Applications of Artificial Intelligence}, vol.~68, pp. 205--213, 2018. [Online]. Available: \url{https://www.sciencedirect.com/science/article/pii/S0952197617302920}
\BIBentrySTDinterwordspacing

\bibitem{PSOTuto}
\BIBentryALTinterwordspacing
F.~Marini and B.~Walczak, ``Particle swarm optimization (pso). a tutorial,'' \emph{Chemometrics and Intelligent Laboratory Systems}, vol. 149, pp. 153--165, 2015. [Online]. Available: \url{https://www.sciencedirect.com/science/article/pii/S0169743915002117}
\BIBentrySTDinterwordspacing

\bibitem{GAN2019107}
\BIBentryALTinterwordspacing
H.~Gan, Y.~Fan, Z.~Luo, R.~Huang, and Z.~Yang, ``Confidence-weighted safe semi-supervised clustering,'' \emph{Engineering Applications of Artificial Intelligence}, vol.~81, pp. 107--116, 2019. [Online]. Available: \url{https://www.sciencedirect.com/science/article/pii/S0952197619300302}
\BIBentrySTDinterwordspacing

\bibitem{SALEHI2021667}
\BIBentryALTinterwordspacing
F.~Salehi, M.~R. Keyvanpour, and A.~Sharifi, ``Smkfc-er: Semi-supervised multiple kernel fuzzy clustering based on entropy and relative entropy,'' \emph{Information Sciences}, vol. 547, pp. 667--688, 2021. [Online]. Available: \url{https://www.sciencedirect.com/science/article/pii/S0020025520308562}
\BIBentrySTDinterwordspacing

\bibitem{PENG2021107683}
\BIBentryALTinterwordspacing
S.~Peng, W.~Ser, B.~Chen, and Z.~Lin, ``Robust semi-supervised nonnegative matrix factorization for image clustering,'' \emph{Pattern Recognition}, vol. 111, p. 107683, 2021. [Online]. Available: \url{https://www.sciencedirect.com/science/article/pii/S0031320320304866}
\BIBentrySTDinterwordspacing

\bibitem{gan2023adaptive}
H.~Gan, Z.~Yang, and R.~Zhou, ``Adaptive safety-aware semi-supervised clustering,'' \emph{Expert Systems with Applications}, vol. 212, p. 118751, 2023.

\bibitem{Wang2023GCFlowAG}
\BIBentryALTinterwordspacing
T.~Wang, F.~Mirzazadeh, X.~Zhang, and J.~Chen, ``Gc-flow: A graph-based flow network for effective clustering,'' \emph{ArXiv}, vol. abs/2305.17284, 2023. [Online]. Available: \url{https://api.semanticscholar.org/CorpusID:258960696}
\BIBentrySTDinterwordspacing

\bibitem{kulis2005semi}
B.~Kulis, S.~Basu, I.~Dhillon, and R.~Mooney, ``Semi-supervised graph clustering: a kernel approach,'' in \emph{Proceedings of the 22nd international conference on machine learning}, 2005, pp. 457--464.

\bibitem{fix1985discriminatory}
E.~Fix, \emph{Discriminatory analysis: nonparametric discrimination, consistency properties}.\hskip 1em plus 0.5em minus 0.4em\relax USAF school of Aviation Medicine, 1985, vol.~1.

\bibitem{cover1967nearest}
T.~Cover and P.~Hart, ``Nearest neighbor pattern classification,'' \emph{IEEE transactions on information theory}, vol.~13, no.~1, pp. 21--27, 1967.

\bibitem{chen2020fast}
Y.~Chen, X.~Hu, W.~Fan, L.~Shen, Z.~Zhang, X.~Liu, J.~Du, H.~Li, Y.~Chen, and H.~Li, ``Fast density peak clustering for large scale data based on knn,'' \emph{Knowledge-Based Systems}, vol. 187, p. 104824, 2020.

\bibitem{abu2019effects}
H.~A. Abu~Alfeilat, A.~B. Hassanat, O.~Lasassmeh, A.~S. Tarawneh, M.~B. Alhasanat, H.~S. Eyal~Salman, and V.~S. Prasath, ``Effects of distance measure choice on k-nearest neighbor classifier performance: a review,'' \emph{Big data}, vol.~7, no.~4, pp. 221--248, 2019.

\bibitem{bora2017performance}
D.~J. Bora, ``Performance comparison of k-means algorithm and fcm algorithm with respect to color image segmentation,'' \emph{International Journal of Emerging Technology and Advanced Engineering}, vol.~7, no.~8, pp. 460--470, 2017.

\bibitem{zhang2016learning}
J.~Zhang, X.~Wu, and V.~S. Sheng, ``Learning from crowdsourced labeled data: a survey,'' \emph{Artificial Intelligence Review}, vol.~46, pp. 543--576, 2016.

\bibitem{desmond2021semi}
M.~Desmond, E.~Duesterwald, K.~Brimijoin, M.~Brachman, and Q.~Pan, ``Semi-automated data labeling,'' in \emph{NeurIPS 2020 Competition and Demonstration Track}.\hskip 1em plus 0.5em minus 0.4em\relax PMLR, 2021, pp. 156--169.

\bibitem{zhang2021survey}
S.~Zhang, O.~Jafari, and P.~Nagarkar, ``A survey on machine learning techniques for auto labeling of video, audio, and text data,'' \emph{arXiv preprint arXiv:2109.03784}, 2021.

\bibitem{misc_liver_disorders_60}
``{Liver Disorders},'' UCI Machine Learning Repository, 1990, {DOI}: https://doi.org/10.24432/C54G67.

\bibitem{misc_dermatology_33}
N.~Ilter and H.~Guvenir, ``{Dermatology},'' UCI Machine Learning Repository, 1998, {DOI}: https://doi.org/10.24432/C5FK5P.

\bibitem{misc_waveform_database_generator_(version_1)_107}
L.~Breiman and C.~Stone, ``{Waveform Database Generator (Version 1)},'' UCI Machine Learning Repository, 1988, {DOI}: https://doi.org/10.24432/C5CS3C.

\bibitem{frank2010uci}
A.~Frank, ``Uci machine learning repository,'' \emph{http://archive. ics. uci. edu/ml}, 2010.

\bibitem{kaggle_diabetes}
\BIBentryALTinterwordspacing
Kaggle. (2024) Kaggle datasets repository. Accessed: 2024-06-08. [Online]. Available: \url{https://www.kaggle.com/datasets/akshaydattatraykhare/diabetes-dataset}
\BIBentrySTDinterwordspacing

\bibitem{misc_heart_disease_45}
A.~Janosi, W.~Steinbrunn, M.~Pfisterer, and R.~Detrano, ``{Heart Disease},'' UCI Machine Learning Repository, 1988, {DOI}: https://doi.org/10.24432/C52P4X.

\bibitem{kaggle_wdbc}
\BIBentryALTinterwordspacing
Kaggle. (2024) Kaggle datasets repository. Accessed: 2024-06-08. [Online]. Available: \url{https://www.kaggle.com/datasets/uciml/breast-cancer-wisconsin-data}
\BIBentrySTDinterwordspacing

\bibitem{kaggle_2024}
\BIBentryALTinterwordspacing
------. (2024) Kaggle: Your home for data science. Accessed: 2024-06-08. [Online]. Available: \url{https://www.kaggle.com}
\BIBentrySTDinterwordspacing

\bibitem{turney1994cost}
P.~D. Turney, ``Cost-sensitive classification: Empirical evaluation of a hybrid genetic decision tree induction algorithm,'' \emph{Journal of artificial intelligence research}, vol.~2, pp. 369--409, 1994.

\bibitem{bishop2006pattern}
C.~M. Bishop and N.~M. Nasrabadi, \emph{Pattern recognition and machine learning}.\hskip 1em plus 0.5em minus 0.4em\relax Springer, 2006, vol.~4, no.~4, chapter 9.

\bibitem{fraley2002model}
C.~Fraley and A.~E. Raftery, ``Model-based clustering, discriminant analysis, and density estimation,'' \emph{Journal of the American statistical Association}, vol.~97, no. 458, pp. 611--631, 2002.

\end{thebibliography}

\vspace{12pt}

\end{document}